\documentclass[10pt, conference,letterpaper]{IEEEtran}
\ifCLASSINFOpdf
\else
\fi

\usepackage[notransparent]{svg}
\usepackage{multirow}
\usepackage{array, makecell}
\usepackage{graphicx} 
\usepackage{subcaption} 
\usepackage{algorithm}
\usepackage{algpseudocode}
\usepackage{amssymb}
\usepackage{mathdots}
\usepackage{yhmath}
\usepackage{hyperref}
\usepackage{amsmath} 
\usepackage[capitalise]{cleveref}
\begin{document}

\title{Learning Online Policies for Person Tracking in Multi-View Environments
}
\author{\IEEEauthorblockN{ Keivan Nalaie}
\IEEEauthorblockA{\textit{Dept. of Computing and Software} \\
\textit{McMaster University}\\
Hamilton, Canada \\
nalaiek@mcmaster.ca}
\and
\IEEEauthorblockN{Rong Zheng}
\IEEEauthorblockA{\textit{Dept. of Computing and Software} \\
\textit{McMaster University}\\
Hamilton, Canada \\
rzheng@mcmaster.ca}

}

\maketitle
        
\begin{abstract}
In this paper, we introduce MVSparse, a novel and efficient framework for cooperative multi-person tracking across multiple synchronized cameras. The MVSparse system is comprised of a carefully orchestrated pipeline, combining edge server-based models with distributed lightweight Reinforcement Learning (RL) agents operating on individual cameras. These RL agents intelligently select informative blocks within each frame based on historical camera data and detection outcomes from neighboring cameras, significantly reducing computational load and communication overhead. The edge server aggregates multiple camera views to perform detection tasks and provides feedback to the individual agents. By projecting inputs from various perspectives onto a common ground plane and applying deep detection models, MVSparse optimally leverages temporal and spatial redundancy in multi-view videos. Notably, our contributions include an empirical analysis of multi-camera pedestrian tracking datasets, the development of a multi-camera, multi-person detection pipeline, and the implementation of MVSparse, yielding impressive results on both open datasets and real-world scenarios. Experimentally, MVSparse accelerates overall inference time by 1.88X and 1.60X compared to a baseline approach while only marginally compromising tracking accuracy by 2.27\% and 3.17\%, respectively, showcasing its promising potential for efficient multi-camera tracking applications.

\end{abstract}

\begin{IEEEkeywords}
Distributed learning, Multi-camera pedestrian tracking, Reinforcement learning, Edge computing.
\end{IEEEkeywords}

\IEEEpeerreviewmaketitle

\section{Introduction}
Multi-object tracking has drawn much interest due to its wide applications in surveillance, search and rescue, and crowd analysis. Compared to single-camera tracking, tracking with multiple synchronized cameras with overlapping fields of view (FoV) has the advantage of better accuracy from less occlusion and multi-view coverage of the same subject. However, the need to fuse information from geographically distributed cameras poses unique challenges. First, since cameras are typically placed at considerable distances from one another in order to reduce costs, there exist substantial variations in perspectives and illumination conditions across different visual fields making object association between identities from different viewpoints difficult. Second, processing multi-view feeds centrally does not scale well. Transferring raw data from multiple cameras to a central server (or a cloud data center) for further processing may incur excessive communication delay. On the other hand, modern object-tracking systems employ a technique called tracking via detection~\cite{zhang2022bytetrack,centerTrack,trackingwithoutshistles,zhang2020fairmot}. In this method, objects are detected and located in a scene using a deep model, and then they are linked based on visual and motion characteristics to construct trajectories. Typical networks for object detection usually contain a large number of convolution layers, and the computational cost of these convolutional neural networks (CNNs) is too high for real-time processing, on resource-limited end devices. Communication among cameras for information fusion also incurs additional delays. 

Fortunately, there exists significant redundancy both spatially and temporally in multi-view videos ~\cite{you2020real,hou2020multiview}. Humans generally move in confined areas such as pathways and sidewalks in outdoor environments. Therefore, a significant proportion of frames contain only static or slowly changing backgrounds. Over time, intermediate features from these regions remain mostly unchanged. We call such redundancy \textit{temporal redundancy}. It is thus wasteful to assign equal amounts of processing to all regions in an input frame. Across cameras, due to their overlapping FoVs, people can be captured by multiple cameras simultaneously, only a subset of which are needed for tracking purposes. Such redundancy is termed \textit{spatial redundancy}.
Existing works on multi-target tracking exploit either temporal or spatial redundancy but rarely both. For example, to leverage temporal redundancy in video analytics in single-camera streams, researchers utilized optical flow to separate foreground moving objects from stationary background~\cite{zhu2017deep,su2023motion}. NetWarp~\cite{gadde2017semantic} warps internal network representations using the estimated optical flow between adjacent frames to accelerate video segmentation. Optical flow incurs extra computing overhead and is inadequate for large motions, such as newly arrived objects. In another line of work, BlockCopy~\cite{verelst2021blockcopy} incorporates a reinforcement learning (RL) model trained online to identify informative image regions from a single camera. Visual features are computed for informative regions only, while those from non-informative ones are ``copied`` from previous frames to achieve computational savings. In order to exploit the spatial redundancy presented in video frames, works in \cite{guo2021crossroi,yang2023novel} perform offline profiling to partition visual fields into sub-regions and assign one camera to each region. The designated camera is thus responsible for object tracking in the corresponding areas. Such approaches can reduce the computational burden on the cameras or decrease the amount of network traffic between cameras and a central server. However, as evident from our preliminary study in Section~\ref{mv_motivation} and existing literature~\cite{hou2020multiview}, the optimal camera for an object is perspective-dependent and changes over time due to movements, lighting, and occlusion. Moreover, fusing multiple views is often critical to maintain high detection and tracking accuracy~\cite{hou2020multiview,you2020real,kohl2020mta}. 

In this paper, we propose MVSparse, an efficient cooperative multi-person tracking framework across multiple synchronized cameras. The MVSparse pipeline consists of models executed on an edge server and distributed lightweight RL agents running on individual cameras that identify the informative blocks in a frame based on past frames on the same camera and detection results from other cameras. Only selected blocks will be sent to the edge server, which is responsible for aggregating multiple views for detection as well as providing feedback to individual agents. The former is accomplished by first projecting inputs from different perspectives to a common ground plane and then applying a deep detection model. The feedback to each agent is obtained from a novel clustering algorithm that associates objects detected by different cameras. The RL agents are trained online in a self-supervised manner so that they can adapt to human movements, scene dynamics, or even camera configuration changes. MVSparse concurrently exploits temporal and spatial redundancy in multi-view videos with small computation and communication overhead. To summarise, the main contributions of this study are as follows:
\begin{itemize}
\item We empirically analyze a multi-camera pedestrian tracking dataset and quantify the degree and dynamics of overlapping views. An Oracle is devised to determine the minimum number of blocks necessary for detection as an objective measure of spatial redundancy. 
\item We propose a multi-camera, multi-person detection pipeline that exploits temporal and spatial redundancy to accelerate inference time and reduce network transfer delay. Our pipeline is lightweight, and coupled with a primary detector, it can be easily adapted to different scenarios. 
\item MVSparse has also been implemented and evaluated using open datasets and in a real-world testbed. Experiment results on the datasets show that it can accelerate the overall inference time by 1.88X and 1.60X compared to a baseline approach that takes all views while marginally degrading tracking accuracy by 2.27\% and 3.17\%, respectively.
\end{itemize}

The rest of the paper is structured as follows:
 We begin with a review of relevant works in Section~\ref{related_work_section} before moving on to a study of cross-camera spatial redundancy for object detection and the effects of motion on camera coverage in multi-view settings in Section~\ref{mv_motivation}.
The proposed methodology is described in Section~\ref{framework}, and experimental findings are discussed in Section~\ref{experiments}. We conclude the paper in Section~\ref{conclusion}.  

\section{Related work} \label{related_work_section} 
In recent years, much research has been done to reduce the heavy computation requirements of deep models in video analytics tasks. Among them, of most relevance to MVSparse are two lines of work: ones that exploit temporal or spatial redundancy in single or multi-camera videos, and ones that enable sparse neural network operations. In both lines of work, we limit the discussion to ones that are directly applicable to multi-target detection and tracking tasks. 

\subsection{Exploration of temporal/spatial redundancy}
Recognizing the abundance of static or slowly changing backgrounds in input frames, BlockCopy was introduced in ~\cite{verelst2021blockcopy} to speed up deep neural network processing for video analytics. It employs a policy network to identify and remove backgrounds and stationary regions from a frame and concentrates exclusively on informative regions. Consequently, only the selected regions in the current frame are processed via sparse convolution, and the computed features of excluded regions are reused from previous frames. BlockCopy has shown promises in accelerating pedestrian detection, instance segmentation, and semantic segmentation tasks, but it is restricted to single-camera inputs. In DeepCache~\cite{xu2018deepcache}, the internal structure of a deep architecture is utilized to produce reusable results by matching regions in the current frame with the earlier frames via diamond searches. Thus, computation is necessary only in mismatched areas in the backbone layers. Similar to BlockCopy, DeepCache only exploits temporal redundancy in single camera feeds. Furthermore, for large input frames, diamond searches can impose significant overhead.

CrossRoI~\cite{guo2021crossroi} is among the first works that exploits spatial redundancy in multi-camera multi-target detection using deep models. It reduces the need for extra communication and computational resources by taking advantage of the overlapping FoVs. Through offline profiling, a lookup table of regional associations among all the cameras is constructed using re-ID filtering. At the inference time, each camera transmits only areas of its captured view specified by the lookup table. Upon receiving camera feeds from multiple perspectives, the server uses sparse block processing, such as SBNet~\cite{ren2018sbnet}, to speed up inference. In contrast to our approach, CrossRoI relies on fixed RoI masks of the visual scenes computed from multiple frames offline. When the density or moving patterns of objects change, this offline approach may fail to provide sufficient coverage to all objects in the scene. Polly~\cite{li2023cross} also exploits spatial redundancy across cameras but allows sharing of detection results in overlapping FoVs among several cameras. Polly eliminates redundant inference over the same regions across different viewpoints by sharing the inference results from the reference camera with the target camera. However, we find that it is beneficial to aggregate different perspectives for object detection and the optimal views change over time. Unlike CrossROI and Polly, MVSparse judiciously adapts the regions in each view to be processed through distributed reinforcement learning and fuses multi-view information to obtain the final detection results. 

Dai \textit{et al.} ~\cite{dai2022respire} developed RESPIRE, a method that aims to reduce spatial and temporal redundancy by carefully choosing frames that maximize overall information given latency, computation resources, and bandwidth constraints. This approach is restricted to general feature descriptors such as SIFT, which results in low granularity when number of objects is large. Also, all camera feeds must be retrieved to fully benefit from any temporal or spatial redundancy.
In~\cite{yang2023novel}, Yang \textit{et al.} proposed a traffic-related object detection framework CEVAS, which simultaneously eliminates the existing spatial and temporal redundancy in multi-view video data. It reduces temporal redundancy by detecting newly arrived objects using optical flow as a motion feature. Targeting spatial redundancy, an object manager runs centrally to evaluate detections across multiple cameras to eliminate spatial redundancy of common objects among different cameras with overlapping FoVs. Compared to these works, MVSparse exploits temporal and spatial redundancy using online reinforcement learning and clustering approaches from beginning to end in an online cooperative fashion.

\subsection{Sparse processing in deep models}
Generally, highly complex deep networks are needed to achieve the state of the art accuracy on visual tasks. However, it may not be necessary to process all layers of a network and instead adjust the network's depth according to input characteristics\cite{veit2018convolutional}. Works such as ~\cite{wu2018blockdrop} and ~\cite{wang2018skipnet} utilize a set of residual convolution layers, and determine the depth of the network sequentially based on past decisions. Instead of applying convolutional filters to the entire input frames, some works only convolve these filters over particular locations~\cite{Chen_2022_CVPR,gao2022convmae}. For example, SBNet~\cite{ren2018sbnet} computes block-wise convolutions based on binary computation masks and copies the results to the corresponding coordinates. Authors in \cite{verelst2020segblocks} introduced SegBlocks, which divides images into blocks and processes low-complexity areas at a reduced resolution to capitalize on spatial redundancy in images. It first splits an image into blocks, runs a policy network to identify non-important regions, and then processes non-important regions at a low resolution. SegBlocks enables feature propagation using zero padding to eliminate discontinuities at block borders.

\section{A preliminary study on multi-camera multi-target pedestrian tracking}
\label{mv_motivation}
Object identification, feature extraction, and object association are the three core components of DNN-based trackers. Reference \cite{nalaie2022deepscale} suggests that the most time-consuming step in the tracking process is object detection. Thus, investigating the computing costs of object detection (amount of areas processed for each frame), specifically in relation to the extent of overlapping areas across many cameras, is the primary goal of the section. We consider the WildTrack dataset~\cite{chavdarova2018wildtrack} which was gathered from 7 static synchronized and calibrated cameras, with overlapping FoVs in outdoor areas. There are 20 people on average per frame.

\begin{figure}[t]
\centerline{\includegraphics[width=0.9\columnwidth]{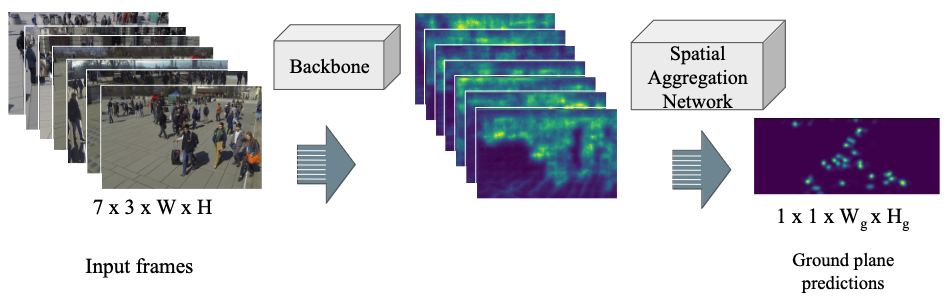}}
\caption[MVDet system.]{MVDet for object detection from multi-camera views. 
}
\label{fig:mv_org_pipeline}
\vspace{-1.2em}
\end{figure}

The model architecture for the baseline multi-view object detector (MVDet~\cite{hou2020multiview}) is shown in Figure~\ref{fig:mv_org_pipeline}. The model collects frames from multiple viewpoints, extracts intermediate features using a deep backbone separately, and then aggregates the multi-view features for ground plane predictions.

\begin{figure}[b]
\centering
\vspace{-1.5em}
\subfloat[Camera coverage.]{\includegraphics[width=0.50\columnwidth]{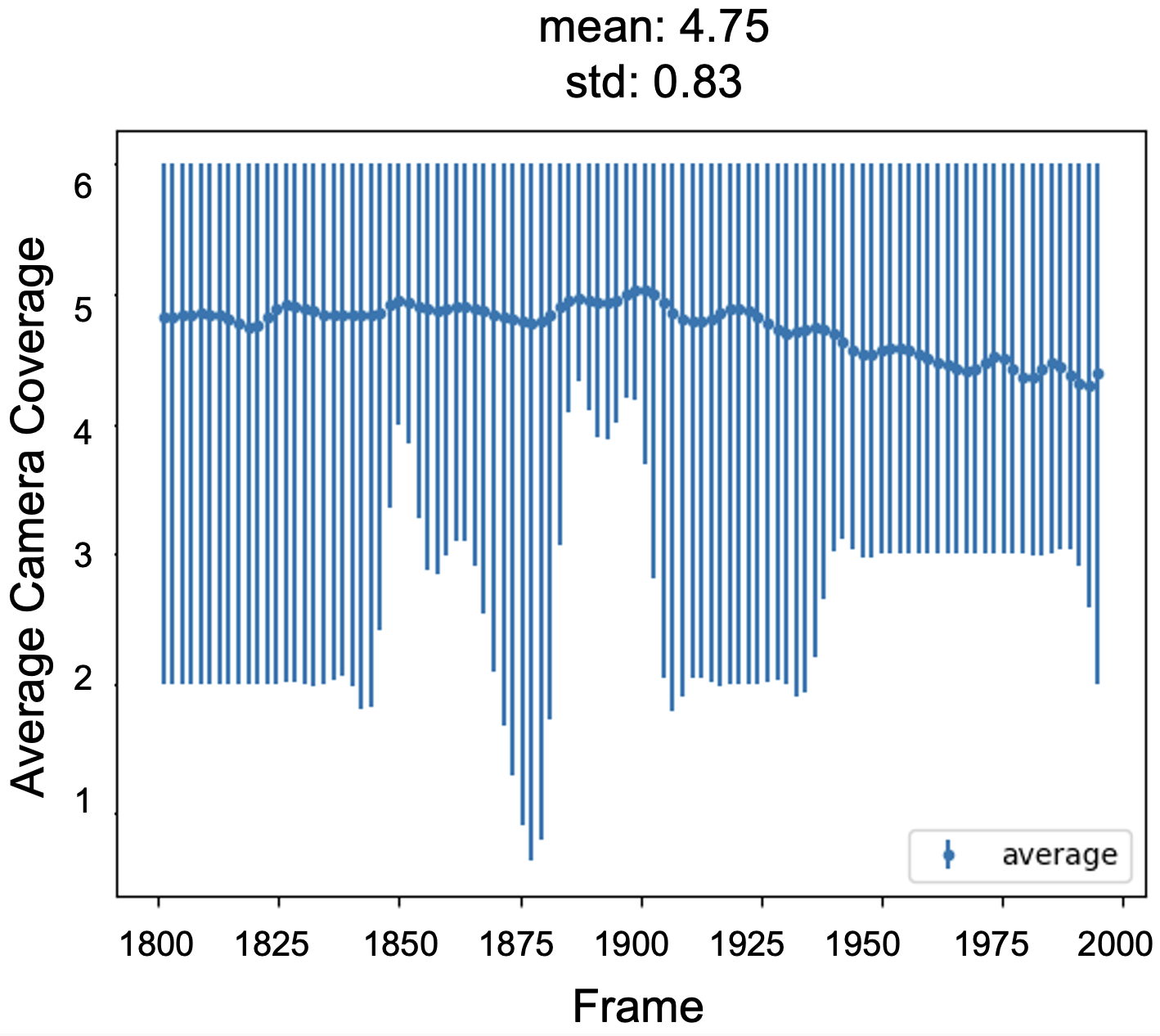}}
\subfloat[Top camera chosen by Oracle.]{ \includegraphics[width=0.50\columnwidth]{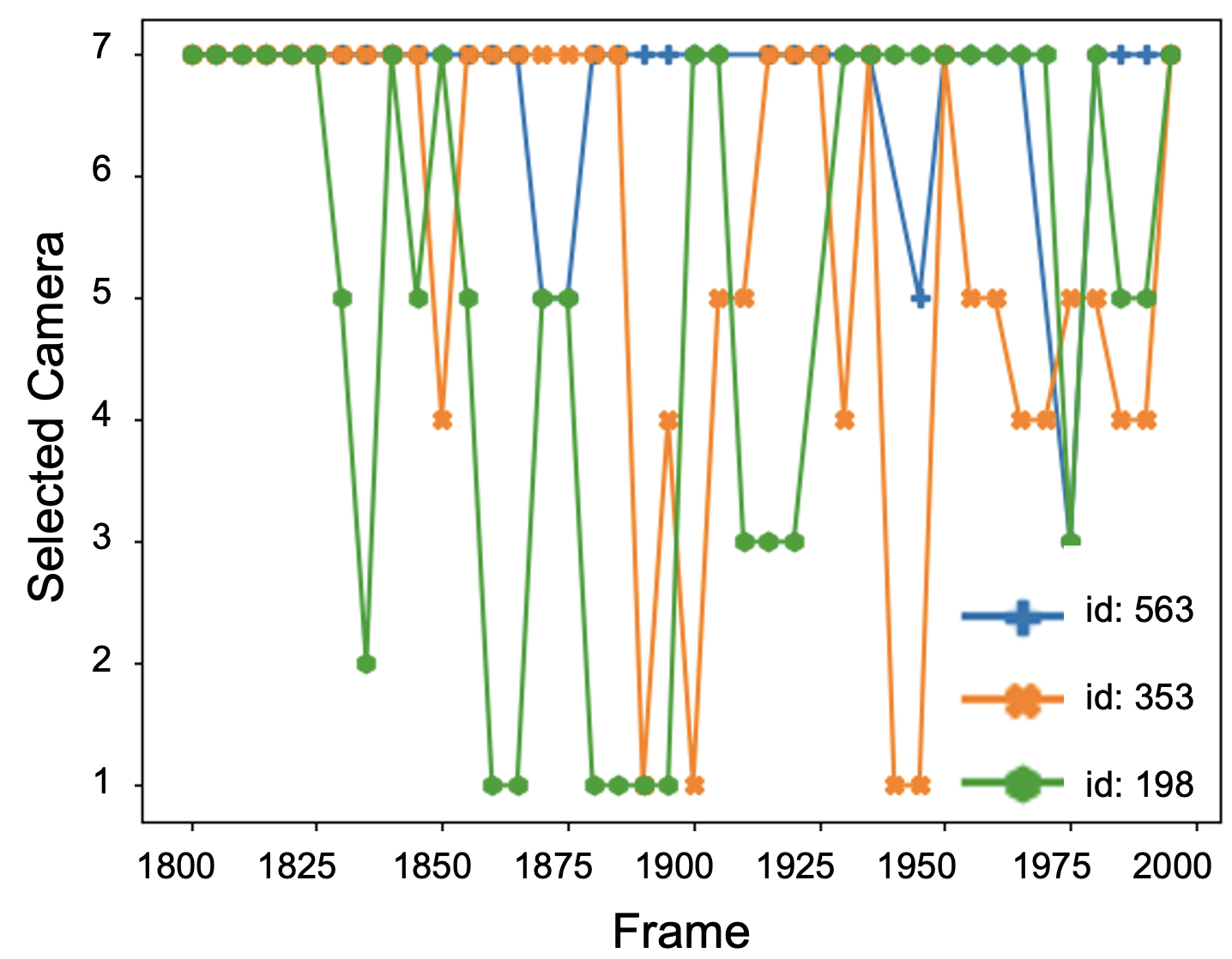}}
\caption{Dynamics in camera coverages and the optimal cameras for different persons.}
\label{fig:presenting_overlaps_coverage}
\end{figure}

\paragraph*{Camera Coverage}
Figure~\ref{fig:presenting_overlaps_coverage}.a shows the minimum, maximum and average number of cameras that can detect a specific person in each frame in WildTrack, based on ground truth annotations. As evident from the figure, there is significant overlap (and consequently spatial redundancy) amongst the camera views. In fact, each person is on average detectable by 4.75 cameras. Also, it can be observed that the degree of overlapping changes over time due to the movements of people. 

\paragraph*{Lower bound on informative regions}
Next, we investigate among the overlapping views, the minimum amount of informative regions necessary to achieve a detection performance comparable to a baseline that takes all views. 

Following BlockCopy \cite{verelst2021blockcopy}, the input frames are first divided into blocks of $128\times 128$px as basic processing units. Thus, there are in total $5\times 9$ blocks in input images of size $640\times 1152$px in WildTrack dataset from seven cameras. To approximate the least number of blocks necessary, we devise an Oracle that has access to both ground truth target locations and ground plane projections of the outputs from the MVDet backbone for each camera. Oracle chooses the top-K detections from all cameras for each target that are closest to their respective ground truth location using bipartite matching.

 From Table~\ref{table:mvdet_vs_oracle}, it can be seen that when $K = 1$, Oracle can drastically reduce the number of processed blocks per frame with reasonable compromising detection performance. With larger $K$s, as expected, detection performance can be further improved with more camera views at the expense of increased number of processed blocks. Figure~\ref{fig:presenting_overlaps_coverage}.b shows the best camera for detecting a person, defined as the view containing the largest detected bounding box for the subject in the ground truth annotation. It can be seen that the best camera changes quite frequently. This can be attributed to a combination of factors such as the varying distance to each camera and (partial) occlusion by other people, etc. Thus, a fixed partition of the visual field likely yields suboptimal decisions over time.

These empirical results indicate time-varying spatial redundancies among various cameras. This study aims to exploit these redundancies to accelerate object detection and tracking pipelines. To this end, we devise the following strategy: to facilitate temporal feature propagation and sparse convolutions on each camera, we take inspiration from BlockCopy~\cite{verelst2021blockcopy}. However, unlike BlockCopy that only handles single camera inputs, the informative blocks in each view are determined {\it jointly} by accounting for overlapping FoVs across cameras. The main challenge is to design a scalable approach that minimizes the amount of information exchanged between cameras and the central server.

\begin{table}[t]
\footnotesize
\centering
\caption{The average number of processed blocks in Oracle and MVDet.}
 \begin{tabular}{c|c|c|c}
 \hline
  Dataset& Method & \makecell{Processed Blocks\\per Frame $\downarrow$} & MODA\% $\uparrow$\\
\hline
 \multirow{4}{*}{WildTrack}
  & Oracle-top1  & 08.63 & 84.80 \\
  & Oracle-top2  & 12.12 & 87.30 \\
  & Oracle-top3  & 13.76 & 88.00 \\
  & MVDet~\cite{hou2020multiview} & 45.00 & 88.20 \\  
 \hline
\end{tabular}
\label{table:mvdet_vs_oracle}
\vspace{-1.5em}
\end{table}


\section{The MVSaprse framework}
\label{framework}
\begin{figure*}[]
\centerline{\includegraphics[width=1.5\columnwidth]{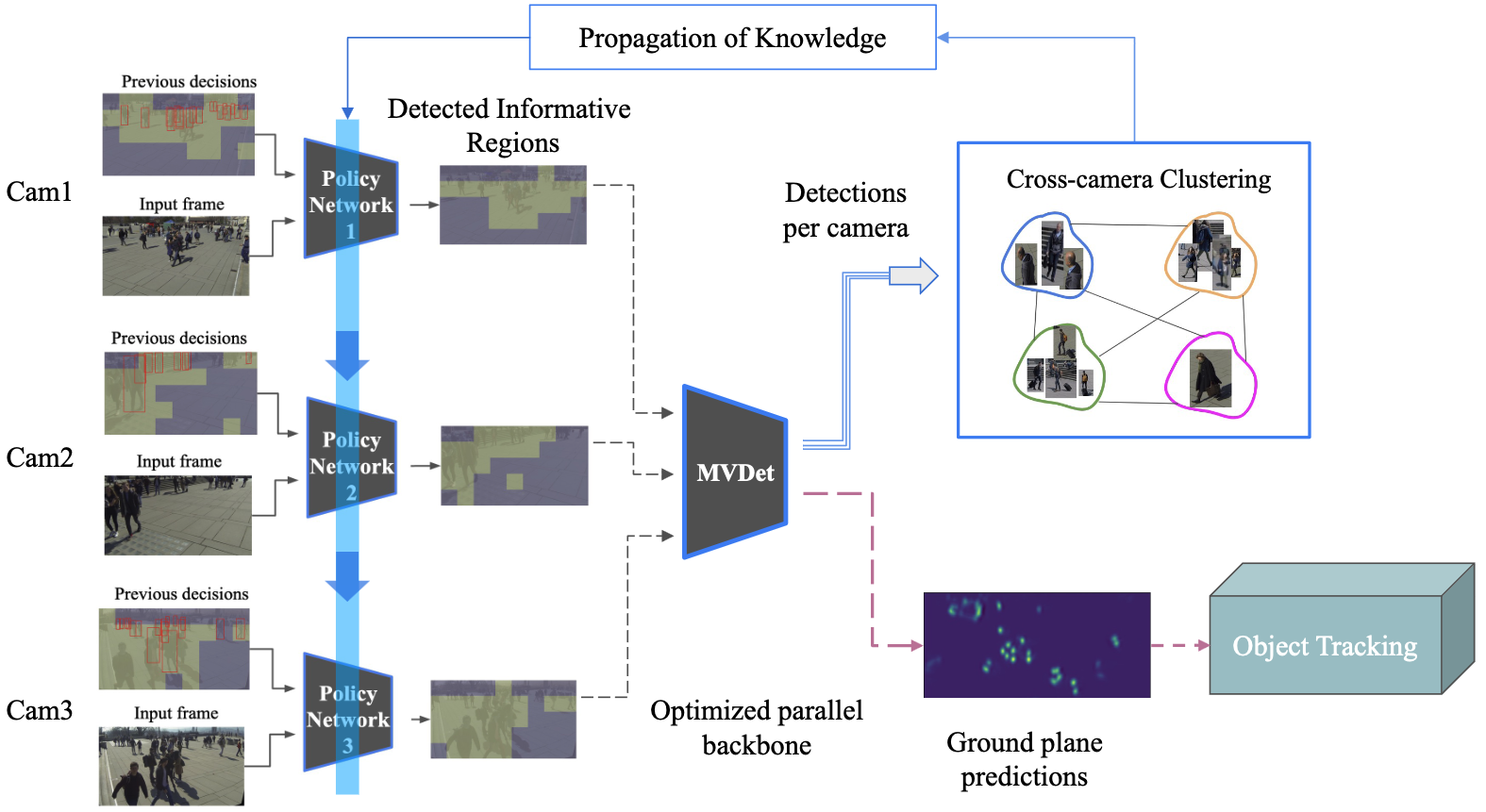}}
\caption[System pipeline of MVSparse]{The system pipeline of MVSparse.} 
\label{fig:mv_pipeline}
\vspace{-1em}
\end{figure*}

The system architecture of MVSparse is shown in Figure~\ref{fig:mv_pipeline}. It consists of a policy network running on each camera that determines the informative blocks in every frame, based on past decisions and the current frame, the MVDet backbone that produces per frame per camera feature maps as well as aggregated ground plane predictions across multiple views, a lightweight tracking module, and a cross-camera clustering module whose outputs are incorporated in the reward functions to train the policy network on-the-fly. Next, we present details of each component in the pipeline. 

\subsection{Multi-view pedestrian detection}
MVSparse takes multiple RGB frames, from different viewpoints, and estimates a pedestrian occupancy map on the ground plane. For each frame in each view, similar to MVDet, it computes $512$ feature maps of size $W_f\times H_f$ using a shared deep backbone. The camera's intrinsic and extrinsic parameters are applied to project these feature maps onto a common ground plane. To aggregate all views, we apply a 3-layer sub-network (called Spatial Aggregation Network) to merge the transformed intermediate features from different viewpoints and generate the final occupancy map on the ground plane. To determine the bounding box of each individual in each view, a sub-network is added to the generated features from each view. Instead of performing bound box regression to predict the center, height, and width of the box containing a person, detection is carried out on a backbone network using head and foot pairing~\cite{hou2020multiview}. The bounding boxes are then used in the clustering algorithm for object association (Section~\ref{section_cross_camera_wise_clustering}) and in tracking.

There are two key differences in the way detection is performed in MVSparse compared to MVDet. First, instead of feeding the entire frames to the detection backbone network, we only update features of informative regions as decided by a policy network (Section~\ref{sparse_policy_section}). Sparse convolutions with high efficiency are not supported by standard deep learning packages like PyTorch. We use SegBlocks~\cite{verelst2020segblocks}, a block-based image processing framework to overcome this limitation. This method converts an RGB image of dimensions of $3\times W \times H$ into blocks of size $ B \times B$. In this paper, we set $B=128$. During execution, only the blocks that require updates are processed, while the representations of the remaining blocks from previous frames are stored and reused using specialized CUDA operations. Second, to further accelerate computation, informative blocks from different views are grouped and processed in parallel. 

\subsection{Cross-camera object association and camera assignment}
\label{section_cross_camera_wise_clustering}
As demonstrated in Section~\ref{mv_motivation}, there exists significant overlaps among the FoVs of different cameras, or in other words, one person can be seen by multiple cameras. This gives rise to potential computation saving by restricting to a small set of views. To do so, two sub-problems should be resolved first, namely, \textit{cross-camera object association} and \textit{camera assignment}. Given a set of detected objects in each view, cross-camera object association aims to group together ones belonging to the same identity. For each distinct identity, camera assignment determines the specific set of $K$ cameras responsible for tracking it, where $K$ is a system parameter.  

The cross-camera object association problem can be formulated as a graph clustering problem. Specifically, let $D^t_c$ be the set of detected bounding boxes of objects in view $c$ at time $t$\footnote{Here, we use the projected coordinates of a bounding box center on the ground plane, its width, and height in the original view to represent each detection.}:
\begin{equation}
    D^t_c = \{d^t_{c,i}: [x_{c,i},y_{c,i},w_{c,i},h_{c,i}],i=1:|D^t_c|\}.
\end{equation}

We define a graph $G(U, E)$, where vertex $u$ stands for an object in a view and an edge exists between two vertex $u_1$ and $u_2$ if and only if they are not detectable by the same camera. The edge weight $w(u_1, u_2)$ is proportional to the Euclidean distance between the respective centers in a ground plane. Thus, the purpose of object association is to partition $G$ to fully connected subgraphs (cliques) such that the sum edge weight in the subgraphs and the number of subgraphs is minimized.  Graph clustering problems are known to be NP-hard~\cite{garey1979computers}. We design a heuristic solution outlined in Algorithm ~\ref{alg:cross_cls_camera_alg}.
\begin{algorithm}
\caption{Clustering detections across multiple cameras}
\label{alg:cross_cls_camera_alg}
\small
\begin{algorithmic}
\Require receives $D^t_c,c\in [1,C]$
\Require outputs set $Clusters^t$ 

\State $Clusters^t=\{D^t_1\}$
\For{\texttt{<View c $\in$ [2:C]>}}
\State $centers \gets Center(Clusters^t)$ 
\State $gpCenters \gets \Pi_g(centers)$. 
\State //projecting into the ground plane
\State $gpD \gets \Pi_g(D^t_c)$. 
\State //projecting into the ground plane
\State $m,um \gets BIP(dist(gpCenters,gpD))$
\For{\texttt{<$(p,q)\gets \texttt{matched m}$>}}
\State $Clusters^t_{p} \gets Clusters^t_{p}\bigcup D^t_{c,q}$
\EndFor
\For{\texttt{<$(u)\gets \texttt{unMatched um}$>}}
\State $Clusters^t\gets Clusters^t\bigcup D^t_{c,u}$
\EndFor
\EndFor
\State \Return $Clusters^t$
\end{algorithmic}
\end{algorithm}

Algorithm~\ref{alg:cross_cls_camera_alg} initializes the clusters using the objects detected in the first view $D^t_1$ and iterates through the remaining views one by one. The ground plane projection function $\Pi_g(\cdot)$ uses camera parameters to project camera coordinates into ground-level coordinates allowing for an associating of clusters with detections based on their respective ground plane distance. For the $c$ th view, we compute the center of each cluster obtained so far and construct a bi-partite graph using Binary Integer Program (BIP) from the cluster centers in one set and the objects in $D^t_{c+1}$ in the other set. An edge exists between a cluster center and an object if their Euclidean distance is below a pre-defined threshold $\epsilon$. Matched objects are included in the respective clusters while the unmatched ones each form a new cluster of size 1. The procedure continues until all views have been considered. 
Therefore, the objects associated with the same person are grouped in one cluster. To determine which cameras are used to detect a person, we select the $K$ largest bounding-box elements from each cluster and only the image blocks in the corresponding views (cameras) are considered informative. This enables us to perform object detection with only $K$ views of the same object in the scene. If $K$ is set to 1, only one view for each identity (cluster) will be chosen. In contrast, when $K = C$, all views should be processed for frame $t$.

Lastly, we define a binary mask $\Gamma^t_c\in \{0,1\}^{M\times N}$, where $M\times N$ are number of blocks at frame $t$. The elements in $\Gamma^t_c$ are set to 1 if the corresponding blocks overlap with a bounding box associated with one of the top-K detections observed in the camera view $c$. Otherwise, they are assigned a value of zero. Also, we define $\gamma^t_c$ as the set of top-K detections identified in camera view $c$. In the subsequent section, $\Gamma^t_c$ and $\gamma^t_c$ are used to determine the information gain and computation costs of camera view $c$.

\subsection{Reinforcement learning for camera-wise sparse processing}
\label{sparse_policy_section}
In the previous section, we showed how detected objects are associated across different views and by selecting only a subset of views for each identity, it is possible to reduce spatial redundancy. However, the question of determining which regions a camera should focus on remains unresolved. This decision should take into account not only the camera's own observed temporal information but also the global spatial information derived from camera assignments. In MVSparse, cameras individually learn over time and decide on the informative blocks based on their local information and ``soft`` global feedback in the form of rewards. This is accomplished through the online training and inference of an RL agent on each camera (Figure~\ref{fig:mv_pipeline}). As it learns, the agent outputs actions such as computing for new features or duplicating previous features for each block and receives a reward depending on the computed information gain and costs. 

\begin{figure}[b]
\vspace{-1.5em}
\centerline{\includegraphics[width=0.7\columnwidth]{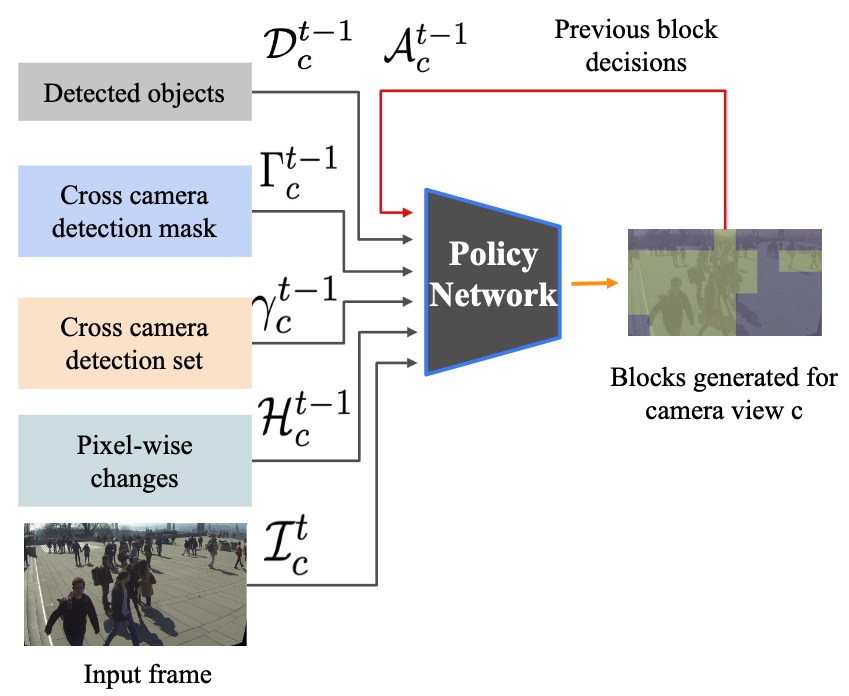}}
\caption[MVSparse's Policy network architecture.]{Policy network on camera $c$.}
\label{fig:policy_net}
\end{figure}

\textbf{Policy Network}: We adopt a model-free RL approach that directly outputs decisions using a policy network. The policy network makes a decision for each block, as illustrated in Figure~\ref{fig:policy_net}. It is designed to be more compact than the primary detection network.  For frame $t$, there are six inputs to this network: the current-frame ($\mathcal{I}^t_c$), pixel-wise changes between frames $t$ and $t-1$ ($\mathcal{H}^{t-1}_c$), cross-camera detection set ($\gamma^{t-1}_c$), cross-camera detection mask ($\Gamma^{t-1}_c$), detected bounding boxes ($\mathcal{D}^{t-1}_c$), and the previous decisions of the policy network ($\mathcal{A}^{t-1}_c$). From these inputs, the policy network outputs $\Psi_t \in [0,1]^{M\times N}$, the probabilities for each of the $M\times N$ blocks at frame $t$. Formally, we have :
\begin{equation}
\begin{split}
    S^t_c = [\mathcal{I}^t_{c},\mathcal{H}^{t-1}_c,\gamma^{t-1}_c,\Gamma^{t-1}_c,\mathcal{D}^{t-1}_c,\mathcal{A}^{t-1}_c]\\
    Policy(S^t_c,\theta_c^t)\rightarrow \Psi^t_{c}\in [0,1]^{M\times N},
    \end{split}
\end{equation}
where subscript $c\in[1:C]$ denotes the camera index, $\theta^t_c$ is a set of network parameters at frame $t$, $S^t_c$ is an input state for camera $c$. Actions $\mathcal{A}^t_c$ is a set of actions for each individual block $a^t_{b,c}$ which is generated by sampling $\Psi^t_c$ according to the Bernoulli distributions to produce binary decisions:
\begin{equation}
    \mathcal{A}^t_c = P_{Bernoulli}(\Psi^t_c)\in \{0,1\}^{M\times N}.
\end{equation}

For $a^t_{b,c}\in \mathcal{A}^t_c$, if $a^t_{b,c}=0$, the features are duplicated from the previous executions, however, when $a^t_{b,c}=1$ the respective block is executed by the detection backbone. 

\paragraph*{Online learning} A policy network deployed in camera $c$ is designed to maximize the following objective function:
\begin{equation}
\label{max_j_theta}
    \max \mathcal{J}(\theta^t_c) = \max \mathbb{E}_{\mathcal{A}^t_c\sim \pi (\theta^t_c)}[\mathcal{R}(\mathcal{A}^t_c)],
\end{equation}
where the reward $\mathcal{R}$ of actions $\mathcal{A}^t_c$ is defined as:
\begin{equation}
    \mathcal{R} (\mathcal{A}^t_c) = \frac{1}{M\times N}\sum_{b=1}^{M\times N}\mathcal{R}(a^t_{b,c}).
\end{equation}

The policy network is updated online using the computed gradient and learning rate $\alpha$:
\begin{equation}
    \theta_{c}^{t+1} \leftarrow \theta_{c}^t + \alpha \nabla_{\theta_c^t}[\mathcal{J}(\theta_c^t)],
\end{equation}

where $\nabla_{\theta_c^t}[\mathcal{J}(\theta_c)]$ can be calculated as:
\begin{equation}
\nabla_{\theta_c^t}[\mathcal{J}(\theta_c^t)] = \nabla_{\theta_c^t} \sum_{b=1}^{M\times N} (\mathbb{E}_{a^t_{b,c}\sim \pi(\theta^t_c)}[\mathcal{R}(a^t_{b,c})]).
\end{equation}

In \cite{verelst2021blockcopy} it is shown that maximizing Eq.(\ref{max_j_theta}) is equivalent to minimizing the following loss function:
\begin{equation}
\label{log_probality_loss}
    \mathcal{L}^t_c= -\sum_{b=1}^{M\times N}\mathcal{R}(a^t_{b,c}) log_{\pi (\theta^t_{c})}(a^t_{b,c}|S^t_{c}),
\end{equation}
where $log_{\pi(\theta^t_{c})}(a^t_{b,c}|S^t_{c})$ is the log probability of action $a^t_{b,c}$ given input state $S^t_c$.

Here, $\mathcal{R}(a^t_{b,c})$ is the reward function w.r.t action $a^t_{b,c}\in \mathcal{A}^t_c$ on camera $c$ at frame $t$ and it is defined as:
\begin{equation}
\label{reinforcement_eq}
    \mathcal{R}(a^t_{b,c}) = 
    \begin{cases}
    \mathcal{R}_{IG}(a^t_{b,c})+ Cost(a^t_{b,c}) \quad a^t_{b,c}=1,\\
    -\mathcal{R}_{IG}(a^t_{b,c})-Cost(a^t_{b,c}) \quad a^t_{b,c}=0.
    \end{cases}
\end{equation}

In Eq.(\ref{reinforcement_eq}), $\mathcal{R}_{IG}$ is the information gain and $Cost$ denotes the computational expense associated with $a^t_{b,c}$. When block $b$ is already activated (or equivalently, $a_{b,c}^t = 1$), it receives a positive reward; otherwise, it receives a negative reward aiming to maximize the objective function in Eq.(\ref{max_j_theta}).

\textbf{Multi-view information gain} The information gain of block $b$ from camera $c$ in the multi-view setting depends on two factors: 1) whether block $b$ is assigned to the camera (based on the algorithm in Section~\ref{section_cross_camera_wise_clustering}), and 2) whether block $b$ provides novel information temporally. (1) is characterized by the $M\times N$ matrix $\Gamma_{c, t}$. For Eq.(\ref{reinforcement_eq}),  we follow the approach in \cite{verelst2021blockcopy} to determine the single-view information gain $IG(a^t_{b,c})$. The procedure eliminates background regions that are stationary while objects that were in motion in the previous frame and do not match entirely or partially with any object in the current frame are associated with higher information gain. Combining (1) and (2), we have:
\begin{equation}
    \mathcal{R}_{IG}(a^t_{b,c})= {IG}(a^t_{b,c}) \Gamma^{t}_{b,c}
\end{equation}
Therefore, the policy network deployed on camera $c$ learns to focus on non-overlapping regions exclusively recognized for viewpoint $c$ by the clustering algorithm.

\textbf{Multi-view computation cost}
In frame $t$, the percentage of processed blocks in viewpoint $c$ is computed as :
\begin{equation}
    P^t_c= \frac{1}{M\times N}\sum_{b=1}^{M\times N} a^t_{b,c}.
\end{equation}
Same as \cite{verelst2021blockcopy}, we compute the moving average of the processed blocks with momentum $\mu$:
\begin{equation}
    \mathcal{M}^t_{c} = (1-\mu)P^t_{c} + \mu P^{t-1}_{c}.
\end{equation}
In the experiments, we set $\mu=0.9$. Then the corresponding computation cost is computed as:
\begin{equation}
\label{r_cost}
    Cost(a^t_{b,c})=  (\tau^t_{c}-\mathcal{M}^t_{c})|\tau^t_c-\mathcal{M}^t_{c}|,
\end{equation}
where $\tau^t_{c}$ is the normalized target cost for a particular view $c$ by considering the ratio between the number of the selected objects in the corresponding view by the clustering algorithm and the maximum number of selected objects among all views:
\begin{equation}
    \tau^t_c = \frac{|\gamma^{t}_c|}{max_{v}|\gamma^{t}_v|},
\end{equation}
where $|\gamma^{t}_c|$ counts the number of bounding boxes in $\gamma^{t}_c$. Eq.(\ref{r_cost}) determines the computation cost based on the amount of informative regions in view $c$. In other words, the computation cost is proportional to the difference between $\tau^t_{c}$ and $\mathcal{M}^t_{c}$. When $a^t_{b,c}=1$ and the percentage of processed blocks $\mathcal{M}^t_{c}$ is under the target $\tau^t_c$, the agent receives a positive reward, leading to lower the loss function, Eq.(\ref{log_probality_loss}); otherwise, the reward is negative and the agent aims to lower the cost. A similar explanation applies when $a^t_{b,c}$ is zero.

\subsection{Lightweight people tracker}
From the identities detected from multiple views, people tracking can be formulated as a path-following problem to connect inferred trajectories from the previous time steps to detections in the present frame. Specifically, in the ground plane, we match an object to an existing trajectory using the IoU criterion. If the computed IoU is higher than a predefined threshold, the trajectory is extended; otherwise, a new trajectory is initiated. The location state of each trajectory is updated in the current frame using a Kalman filter ~\cite{jde}.

\section{Performance evaluation}
\label{experiments}

\subsection{Datasets}
Two multi-view datasets are used in the experiments. The WildTrack dataset contains 400 images of size $1080\times1920$px from 7 cameras encompassing $12\times 36$ square meter area. The ground plane is segmented into cells of size $2.5 cm^2$ with a total of $480\times1440$ cells. On average, $3.74$ cameras are required to cover the entire scene, with approximately 20 persons in each frame.
MultiviewX comprises synthetic images generated by Unity Engine ~\cite{unityEngine} of resolution $1080\times1920$px taken by 6 different cameras. On average, there are 40 people in each frame and 4.41 cameras are needed for full coverage of the scene of size $16\times 25m^2$. The coverage area on the ground plane is divided into $640\times1000$ cells, each of $2.5cm^2$ in size. A total of 400 video frames for each camera are included in the dataset.

Labels (bounding boxes and IDs) are provided for both datasets at 2 FPS. To study the effect of temporal redundancy, a higher frame rate is needed. For WildTrack, the original video sequences at the frame rate 29.85 are used while interpolation is applied in the label space to create bounding boxes in intermediate frames. 
MultiviewX does not contain video sequences at higher FPS that allow the extraction of additional frames. Leveraging FILM~\cite{reda2022film}, we generate new frames at a frame rate of 31.25 by interpolating two successive frames. Bounding boxes in the newly generated frames are obtained through interpolation from those in the original frames in a similar manner as for WildTrack. 
 \begin{figure}[t]
\centering
\subfloat[WildTrack dataset with 7 distinctive views.]{\includegraphics[width=0.80\columnwidth]{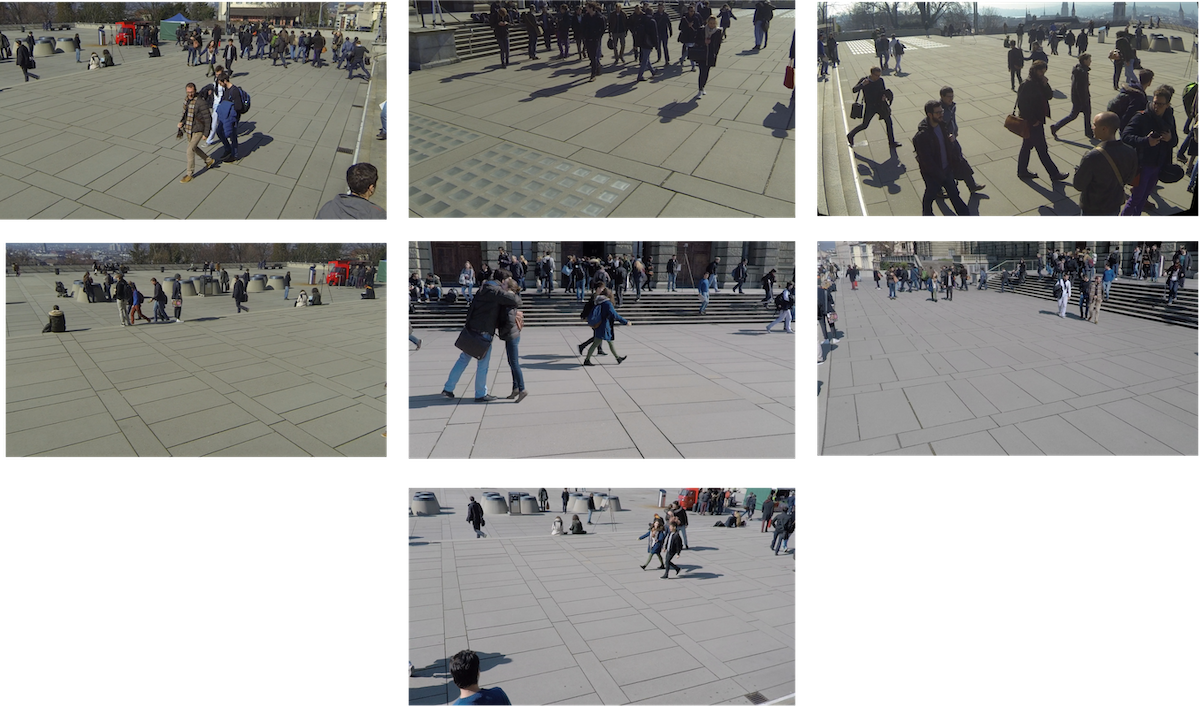}} \\

\subfloat[MultiviewX dataset with 6 different views.]{ \includegraphics[width=0.7\columnwidth]{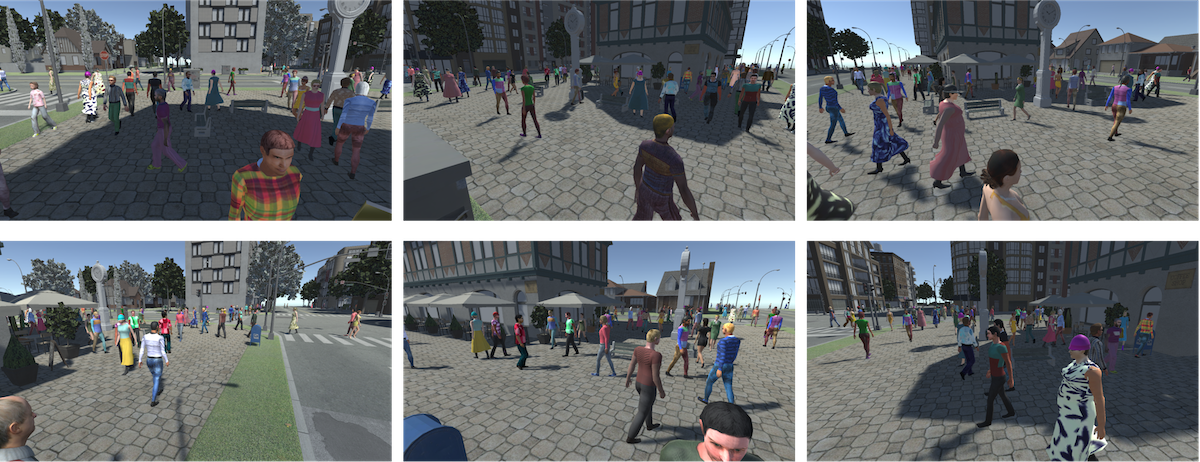}}%
\caption{ An overview of the studied datasets.}
\label{fig:datasets_overview}
\vspace{-1.5em}
\end{figure}

\subsection{Implementation details}
\label{implementation_details}
The frames have been downsampled to $640\times1152$px in order to reduce computation complexity. Each frame is split into a grid of $5\times 9$ blocks, with block size of $128 \times 128$px. The backbone of MVDet uses Resnet-18 to extract features with weights randomly initialized during training. We use a Resnet-8 backbone architecture in the policy network with additional three convolutional layers of 64 channels, batch normalization, and ReLU activations, as well as a softmax layer over the channel dimension. The policy networks are trained every 10 frames. For tracking, we set the conventional threshold for IoU at 0.5, centering a 5-unit square in the location of each detected person in a ground plane.
 
To assess the effectiveness of MVSparse, several metrics are utilized including Average processed blocks per frame, MODA (Multiple Object Detection Accuracy), MODP (Multiple Object Detection Precision ), precision, and recall, for detection as well as tracking metrics MOTA (Multiple Object Tracking Accuracy) and IDF1 (identification F1 score (IDF1), for tracking. 

Specially, MOTA is defined as:  
\begin{equation}
    MOTA = 1 - \frac{\sum_{t} FP_t + FN_t + IDSW_t}{\sum_{t} GT_t},
\end{equation}
where $GT_t$ represents the number of ground truth bounding boxes in frame $t$, $FP_t$, $FN_t$, and $IDSW_t$ are the number of false positive, false negative, and ID switched objects in frame $t$, respectively. $IDF_1$ computes the ratio of correctly identified detections over the number of computed detections and ground truth:
\begin{equation}
    IDF_1 =  \frac{2\times IDTP}{2\times IDTP+IDFP+IDFN} 
\end{equation}
where $IDTP, IDFT,$ and $IDFN$ stand for the number of truly identified objects, and the number of false detections, respectively.

For comparison, we consider three baseline methods: MVDet, BlockCopy, and CrossRoI. We have extended BlockCopy to handle multi-view inputs. Specifically, informative blocks in each view are first processed separately based on the decision of its policy network. The resulting feature maps are then projected to the ground plane and aggregated for final detection. The implementation of  CrossRoI [10] follows the author's released codes in~\cite{CrossRoIimplementation}. This method involves constructing a lookup table by associating bounding boxes in the dataset. Features are only computed from unmasked RoIs in all camera views using SegBlock. For a fair comparison, we have also trained the MVDet backbone on WildTrack and MultiViewX using the masks extracted from CrossRoI. CrossRoI's SVM $\gamma$ and RANSAC parameters for WildTrack and Multiview datasets are set to ($5\times 10^{-6},1.0$) and ($1\times 10^{-6},1.0$), respectively.

 \subsection{Multi-camera detection} 
 \label{multi_cam_detection}
The detection performance of MVSparse and baseline methods are summarized in \cref{table:mv_det_wildtrack_ds}. We set $K=3$ as a default value for MVSparse. MVDet can reach an accuracy of $87.6\%$ ($87.10\%$) in WildTrack (MultiviewX) by processing an entire image. By duplicating non-informative regions and running detection backbone only on informative regions, BlockCopy has comparable accuracy as MVDet but only processes $73\%$ of blocks per frame in both datasets. MVSparse, however, is far more efficient, processing an average of $40\%$ blocks per frame with only a minor reduction in accuracy by $0.7\%$ and $0.4\%$ in WildTrack and MultiViewX respectively.
 
The crowdedness of the scenes in MultiviewX is higher than WildTrack. As a result, CrossRoI finds a higher number of regions and selects more blocks per frame in MultiviewX than WildTrack. However, since the selection is solely based on overlapping criteria without considering detection quality and utilizing multiple views, the MODA scores for CrossRoI in both datasets are lower than those of other methods.

 \begin{table}[t]
\footnotesize
\centering
\caption{Detection results on the WildTrack and MultiviewX datasets.}
\resizebox{\columnwidth}{!}{%
\begin{tabular}{c|c|c|c|c|c|c}
 \hline
  Dataset&Method&\makecell{Processed blocks \\per frame $\downarrow$}&MODA\% $\uparrow$ &MODP\% $\uparrow$& Prec.\% $\uparrow$& Recall\% $\uparrow$\\
 \hline
 \multirow{4}{*}{WildTrack}& 
 MVDet~\cite{hou2020multiview} & 45.00 & 87.60 & 74.80 & 95.80 & 91.50\\
 &BlockCopy~\cite{verelst2021blockcopy} & 33.33 & 87.70 & 74.80 & 95.90 & 91.60\\
 &CrossRoI~\cite{guo2021crossroi} & 22.03 & 77.00 & 73.50 & 90.90 & 85.50 \\
 &MVSparse (ours) & 23.35 & 86.90 & 74.50 & 96.40 & 90.30\\
 \hline
  \multirow{4}{*}{MultiviewX}
 &MVDet~\cite{hou2020multiview} & 45.00 & 87.10 & 80.30 & 98.00 & 88.80\\
 &BlockCopy~\cite{verelst2021blockcopy} & 32.89 & 87.00 & 80.20 & 98.00 & 88.80\\
 &CrossRoI~\cite{guo2021crossroi} & 27.52 & 83.20 & 77.80 & 96.20 & 86.60 \\
 &MVSparse (ours) & 27.57 & 86.70 & 80.20 & 98.00 & 88.80\\
 \hline
\end{tabular}}
\label{table:mv_det_wildtrack_ds}
\vspace{-1.5em}
\end{table}

\begin{figure}[b]
\centering
\vspace{-1.4em}
\subfloat[WildTrack dataset.]{\includegraphics[width=0.25\textwidth]{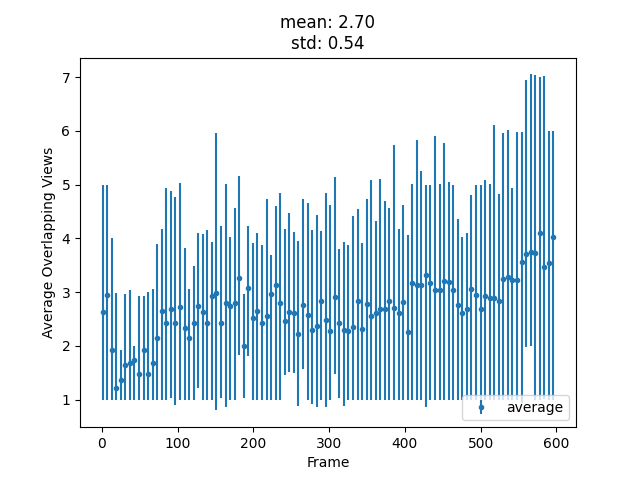}}
\subfloat[MultiviewX dataset.]{\includegraphics[width=0.25\textwidth]{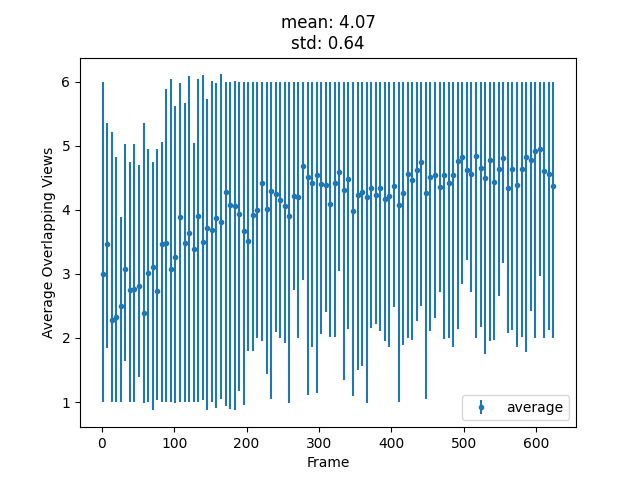}}%
\caption[Average overlapping views selected by MVSparse.]{Average overlapping views selected by MVSparse. Numbers are smoothed using interpolating B-spline~\cite{2020SciPy-NMeth}.}
\label{fig:sparseness}
\end{figure}

Figure~\ref{fig:sparseness} shows the number of views processed per subject in MVSparse. In this experiment, a view is included if the detected bounding box from the respective view overlaps the ground truth box by at least 50\%. When comparing these with Figure.~\ref{fig:presenting_overlaps_coverage}, we observe that MVSparse efficiently exploits spatial redundancy, reducing the average number of views from 4.7 to 2.7 for WildTrack and from 4.87 to 4.07 for MultiviewX.

\subsection{Impact of camera coverage K}
\begin{figure}[t]
\centerline{\includegraphics[width=0.9\columnwidth]{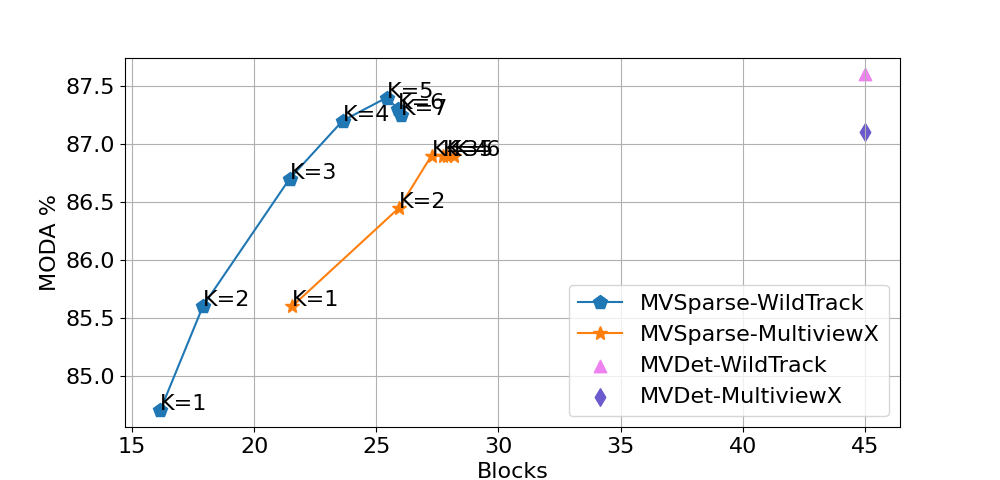}}
\caption{Impact of $K$ on MVSparse detection performance.}
\label{fig:block_moda}
\vspace{-1.5em}
\end{figure}

Recall that the parameter $K$ controls the number of views selected by the clustering algorithm for each identity. Figure~\ref{fig:block_moda} shows the effect of $K$ on the detection accuracy. As expected, as $K$ increases, as more views are aggregated, the detection accuracy increases while the number of blocks processed increases as well. More than 2.5\%  (0.9\%) improvement in MODA can be seen in WildTrack (MultiviewX) when $K$ increases from 1 to 6 at the expense of 60\% (30\%) more processed blocks. According to the results, reasonable accuracy can be attained with fewer views when K is set to 3.


\subsection{Micro-benchmark for parallelization}
\label{micro_benchmark}
\begin{figure}[b]
\vspace{-1.6em}
\centerline{\includegraphics[width=1\columnwidth]{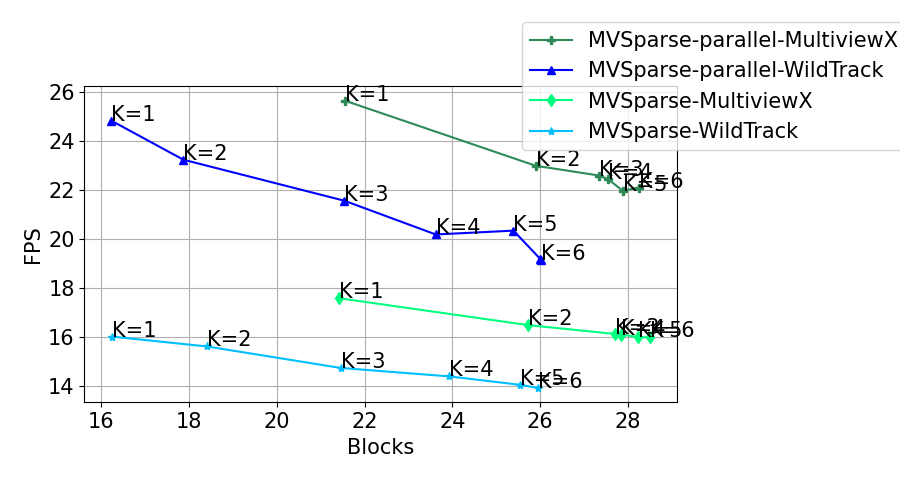}}
\caption{Backbone inference time (FPS) in MVSparse.}
\label{fig:block_fps}
\end{figure}

In this experiment, we study the impact of processing blocks from different views in parallel on MVSparse. Additionally, we calculate the inference time by taking the average processing time (excluding loading and pre-processing operations) on an NVIDIA GTX 3080 10GB GPU with an Intel i7 CPU running Pytorch 1.9, and CUDA 11.6. In Figure~\ref{fig:block_fps}, the backbone inference time (excluding the policy network) with and without parallel processing for different $K$s is shown. Clearly, the inference time decreases as the number of processed blocks decreases in both cases. Moreover, extracting feature maps from blocks from selected views through batch processing further reduces the inference time. 
\begin{table*}[h]
\footnotesize
\centering
\caption{Testbed results on the WildTrack dataset.}
\resizebox{\textwidth}{!}{%
\begin{tabular}{c|c|c|c|c|c|c|c|c}
 \hline
  Method&
  \makecell{Processed Blocks\\per Frame $\downarrow$}&
  FPS $\uparrow$&
  \makecell{Transmission Time (ms)\\per Frame $\downarrow$}&
  \makecell{Server Time (ms)\\ per Frame $\downarrow$}&
  \makecell{Camera Time (ms) \\ per Frame $\downarrow$}&
  \makecell{Network Traffic Load\\MB per Frame $\downarrow$}&
  MODA\% $\uparrow$&MOTA\% $\uparrow$\\
 \hline  
 MVDet~\cite{hou2020multiview} & $45.00\pm0.00$ & $0.67\pm0.03$ & $1426.45\pm67.28$ & $111.475\pm0.52$ & $236.79\pm5.95$ & $2.66\pm0.00$ & $77.50\pm0.00$&$75.00\pm0.00$\\
 BlockCopy~\cite{verelst2021blockcopy} & $33.27\pm0.03$ & $0.77\pm0.03$ & $1148.85\pm48.75$ & $101.75\pm0.00$ & $309.23\pm5.52$ & $1.97\pm0.00$ & $77.66\pm0.04$ & $75.23\pm0.04$ \\
  CrossRoI~\cite{guo2021crossroi} & $ 25.53\pm0.00$  & $ 0.88\pm0.03$ & $ 1149.46\pm 162.40$ & $ 123.64\pm3.45$ & $ 250.58\pm11.37$ & $ 1.51\pm0.00$ & $67.10\pm0.00$ & $ 63.79\pm0.00$\\
 MVSparse (ours) & $14.43\pm0.12$ & $1.26\pm0.02$ & $767.27\pm0.01$ & $77.63\pm0.00$ & $310.13\pm2.47$ & $0.86\pm0.00$ & $75.80\pm0.29$ &$72.83\pm0.03$\\
 \hline
\end{tabular}}
\label{table:mv_jetson_wildtrack_ds}
\end{table*}
 \begin{table*}[h]
\footnotesize
\centering
\caption{Testbed results on the MultiviewX dataset.}
\resizebox{\textwidth}{!}{%
\begin{tabular}{c|c|c|c|c|c|c|c|c}
 \hline
  Method&
  \makecell{Processed Blocks\\per Frame $\downarrow$}&
  FPS $\uparrow$&
  \makecell{Transmission Time (ms)\\per Frame $\downarrow$}&
  \makecell{Server Time (ms)\\ per Frame $\downarrow$}&
  \makecell{Camera Time (ms) \\ per Frame $\downarrow$}&
  \makecell{Network Traffic Load\\MB per Frame $\downarrow$}&
  MODA\% $\uparrow$&MOTA\% $\uparrow$\\
 \hline  
 MVDet~\cite{hou2020multiview} & $45.00 \pm 0.00$ & $0.73\pm 0.02$ & $1188.03\pm48.90$ & $99.87\pm0.11$ & $323.69\pm6.23$ & $2.63\pm0.00$ & $79.70\pm0.00$ &$70.7\pm0.00$\\
 BlockCopy~\cite{verelst2021blockcopy} & $32.86\pm 0.02$ & $0.81\pm 0.00$ & $1043.01\pm20.66$ & $100.05\pm1.03$ & $402.82\pm4.11$ & $1.93\pm0.00$ & $79.80\pm0.00$&$70.8\pm0.00$ \\
 CrossRoI~\cite{guo2021crossroi} & $ 28.52\pm0.00$  & $ 0.85\pm 0.04$ & $ 1035.37\pm154.38$ & $121.76 \pm0.86$ & $ 390.10\pm32.87$ & $ 1.67\pm0.00$ & $64.90\pm0.00$ & $ 58.29\pm0.00$\\
 MVSparse (ours) & $17.46\pm 0.00$ & $1.17\pm0.02$ & $672.14\pm25.60$ & $77.94\pm0.44$ & $399.05\pm8.81$ & $1.03\pm0.00$ & $77.20\pm0.00$ & $67.53\pm0.33$\\
 \hline
\end{tabular}}
\label{table:mv_jetson_multiviewx_ds}
\end{table*}

\subsection{Tracking performance}
 The MOT performance of all methods is presented in Table~\ref{table:mv_tracking_results}. MVSparse achieves a MOTA of $85\%$ on WildTrack, a slight $1.0\%$ decrease in accuracy compared to MVDet by processing $52.33\%$ fewer blocks. Similar observations can be made for MultiviewX. MVSparse processes${\sim}40\%$ fewer blocks per frame, with only $0.1\%$ reduction in MOTA compared to MVDet.
 
\begin{table}[h]
\vspace{-1.5em}
\footnotesize
\centering
\caption{Tracking results on WildTrack and MultiViewX.}
\resizebox{\columnwidth}{!}{%
\begin{tabular}{c|c|c|c|c}
 \hline
  Dataset&Method&\makecell{Processed Blocks\\per Frame $\downarrow$}&MOTA\% $\uparrow$& IDF1\% $\uparrow$\\
 \hline
 \multirow{3}{*}{WildTrack}
 &MVDet~\cite{hou2020multiview} & 45.00 & 86.00 & 84.30\\
 &BlockCopy~\cite{verelst2021blockcopy}  & 33.33 & 86.00 & 83.60\\
 &CrossRoI~\cite{guo2021crossroi} & 22.03 & 75.50 & 73.00 \\
 &MVSparse (ours)   & 21.45  & 85.00 & 82.80\\
  \hline
 \multirow{3}{*}{MultiviewX}
 &MVDet~\cite{hou2020multiview} & 45.00 & 78.00 & 63.30 \\
 &BlockCopy~\cite{verelst2021blockcopy}  & 32.89 & 78.10 & 61.00 \\
 &CrossRoI~\cite{guo2021crossroi} & 27.52 & 74.70 & 55.90 \\
 &MVSparse (ours)   & 27.26  & 77.90 & 60.95 \\
 \hline
\end{tabular}}
\label{table:mv_tracking_results}
\vspace{-1.5em}
\end{table}

\subsection{Testbed implementation and experiments}
\label{mv_testbed}
\begin{figure}[b]
\centering
\centerline{\includegraphics[width=0.79\columnwidth]{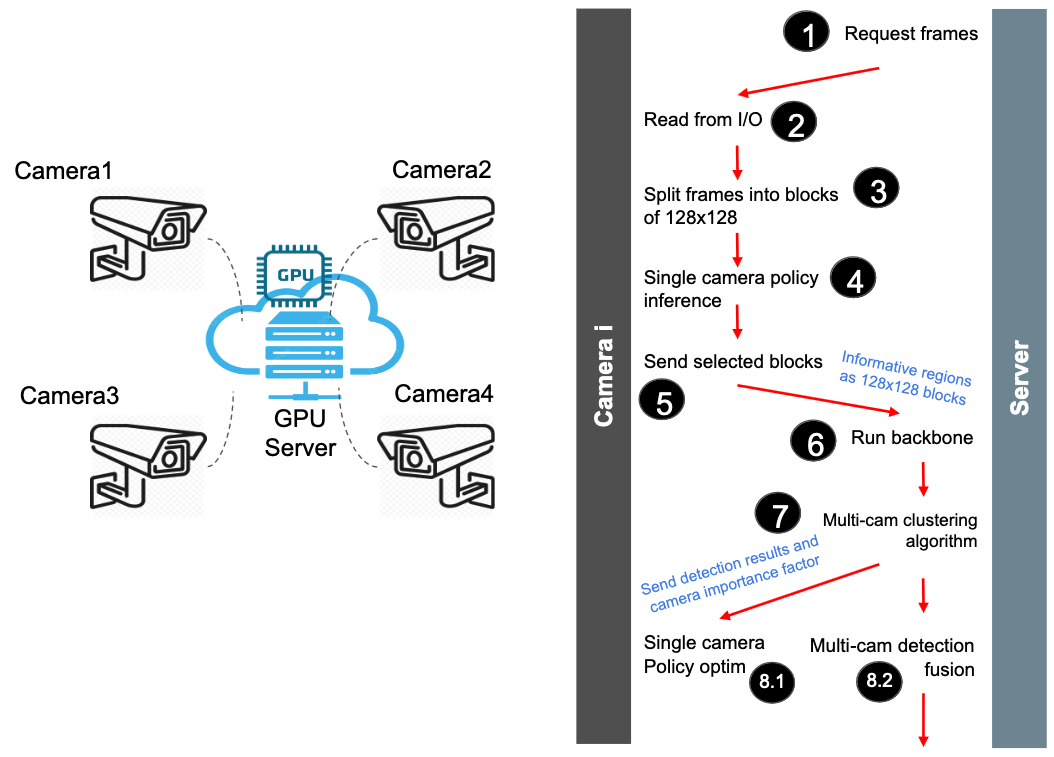}}
\caption{Representative architectures for computation partition between a GPU server and smart cameras in MVSparse.}
\label{fig:testbed_pipelines_mvdet}

\end{figure}

To further evaluate MVSparse's performance in real-world deployment, we have built a small-scale testbed consisting of four NVIDIA Jetson TX2 boards and a desktop computer as described in Section~\ref{micro_benchmark} interconnected via WiFi, with average upload and download bandwidths 18.5 and 21.1 Mbps, respectively. The Jetson embedded GPU boards are used to emulate smart cameras with limited storage and processing capability. In the distributed implementation of MVSparse, the cameras take incoming frames and run the policy network separately. Selected blocks are then transmitted to the server for aggregated detection and tracking. The server also runs the clustering algorithm and sends the results to the respective cameras (Figure~\ref{fig:testbed_pipelines_mvdet}). For BlockCopy, a policy network runs on each camera to select informative blocks to transmit to the server. No coordination is done across cameras. In CrossRoI, each camera only sends blocks according to a fixed lookup table.   
In MVDet, full frames are sent directly from the cameras to the server with no further processing done at the cameras. Each method is trained and tested on the first four camera views of the two datasets. 
 
\cref{table:mv_jetson_wildtrack_ds,table:mv_jetson_multiviewx_ds} compare the accuracy, speed, a breakdown of end-to-end inference time, and the amount of data exchanged between the cameras and the server using different approaches in the testbed. The transmission time is the amount of time needed to transmit data between the cameras and the server. There exists a small timing overlap in camera/ server processing and transmission time due to the multi-threaded implementation. The reported time is averaged over 3 runs of the experiments to mitigate the impact of varying network conditions. It is worth noting that employing only four cameras (out of 6 or 7) for pedestrian detection and tracking reduces the detection accuracy, which explains lower MODA and MOTA scores than those reported in \cref{table:mv_det_wildtrack_ds,table:mv_tracking_results}. Due to extra operations for policy inference, MVSparse spends more time on the camera. However, overhead is compensated by the savings in transmission time because fewer data is transferred from cameras to the server and server processing time. For instance, on WildTrack, compared to MVDet, MVSparse results in a transmission time that is more than twice as quick and only uses $~32\%$ of the network bandwidth. It has a faster server processing time than BlockCopy for two reasons. First, reducing spatial redundancy across several camera viewpoints, MVSparse minimizes the number of blocks required for each frame. Second, blocks from different camera views are processed in parallel through the backbone on the server side. In contrast, BlockCopy processes frames from multiple cameras in consecutive order. Similar to the results in Section \ref{multi_cam_detection}, CrossROI has degraded detection and tracking performance due to the static partition of the scenes among camera views. Furthermore, the number of processed blocks increases due to fewer overlapping among the four cameras in CrossROI.

\section{Conclusion}
\label{conclusion}
This work investigated accelerated tracking of pedestrians in crowded scenes using multiple cameras by exploiting temporal and spatial redundancy. The proposed MVSparse pipeline is fully end-to-end trainable and can be deployed distributedly among cameras and an edge server. Evaluations on open datasets and on a small scale testbed showed that MVSparse could greatly speed up the inference time with only marginal losses in tracking accuracy. Though adopting MVDet as a backbone in the current implementation, MVSparse can work with other object detectors. For future work, we are interested in exploring the potential of MVSparse in other vision tasks such as semantic segmentation and activity recognition in multi-camera settings. 
\bibliographystyle{IEEEtran}
\bibliography{references}

\begin{thebibliography}{10}
\providecommand{\url}[1]{#1}
\csname url@samestyle\endcsname
\providecommand{\newblock}{\relax}
\providecommand{\bibinfo}[2]{#2}
\providecommand{\BIBentrySTDinterwordspacing}{\spaceskip=0pt\relax}
\providecommand{\BIBentryALTinterwordstretchfactor}{4}
\providecommand{\BIBentryALTinterwordspacing}{\spaceskip=\fontdimen2\font plus
\BIBentryALTinterwordstretchfactor\fontdimen3\font minus \fontdimen4\font\relax}
\providecommand{\BIBforeignlanguage}[2]{{%
\expandafter\ifx\csname l@#1\endcsname\relax
\typeout{** WARNING: IEEEtran.bst: No hyphenation pattern has been}%
\typeout{** loaded for the language `#1'. Using the pattern for}%
\typeout{** the default language instead.}%
\else
\language=\csname l@#1\endcsname
\fi
#2}}
\providecommand{\BIBdecl}{\relax}
\BIBdecl

\bibitem{zhang2022bytetrack}
Y.~Zhang, P.~Sun, Y.~Jiang, D.~Yu, F.~Weng, Z.~Yuan, P.~Luo, W.~Liu, and X.~Wang, ``{Bytetrack: Multi-object Tracking by Associating Every Detection Box},'' in \emph{Computer Vision--ECCV 2022: 17th European Conference, Tel Aviv, Israel, October 23--27, 2022, Proceedings, Part XXII}.\hskip 1em plus 0.5em minus 0.4em\relax Springer, 2022, pp. 1--21.

\bibitem{centerTrack}
X.~Zhou, V.~Koltun, and P.~Kr{\"a}henb{\"u}hl, ``{Tracking Objects As Points},'' in \emph{European Conference on Computer Vision}.\hskip 1em plus 0.5em minus 0.4em\relax Springer, 2020, pp. 474--490.

\bibitem{trackingwithoutshistles}
P.~Bergmann, T.~Meinhardt, and L.~Leal-Taixe, ``{Tracking Without Bells And Whistles},'' in \emph{Proceedings of the IEEE/CVF International Conference on Computer Vision}, 2019, pp. 941--951.

\bibitem{zhang2020fairmot}
Y.~Zhang, C.~Wang, X.~Wang, W.~Zeng, and W.~Liu, ``{FairMOT: On The Fairness Of Detection And Re-Identification In Multiple Object Tracking},'' \emph{arXiv preprint arXiv:2004.01888}, 2020.

\bibitem{you2020real}
Q.~You and H.~Jiang, ``{Real-time 3d Deep Multi-camera Tracking},'' \emph{arXiv preprint arXiv:2003.11753}, 2020.

\bibitem{hou2020multiview}
Y.~Hou, L.~Zheng, and S.~Gould, ``Multiview detection with feature perspective transformation,'' in \emph{Computer Vision--ECCV 2020: 16th European Conference, Glasgow, UK, August 23--28, 2020, Proceedings, Part VII 16}.\hskip 1em plus 0.5em minus 0.4em\relax Springer, 2020, pp. 1--18.

\bibitem{zhu2017deep}
X.~Zhu, Y.~Xiong, J.~Dai, L.~Yuan, and Y.~Wei, ``{Deep Feature Flow for Video Recognition},'' in \emph{Proceedings of the IEEE conference on computer vision and pattern recognition}, 2017, pp. 2349--2358.

\bibitem{su2023motion}
J.~Su, R.~Yin, S.~Zhang, and J.~Luo, ``{Motion-state Alignment for Video Semantic Segmentation},'' \emph{arXiv preprint arXiv:2304.08820}, 2023.

\bibitem{gadde2017semantic}
R.~Gadde, V.~Jampani, and P.~V. Gehler, ``{Semantic Video CNNs Through Representation Warping},'' in \emph{Proceedings of the IEEE International Conference on Computer Vision}, 2017, pp. 4453--4462.

\bibitem{verelst2021blockcopy}
T.~Verelst and T.~Tuytelaars, ``{BlockCopy: High-resolution Video Processing with Block-sparse Feature Propagation and Online Policies},'' in \emph{Proceedings of the IEEE/CVF International Conference on Computer Vision}, 2021, pp. 5158--5167.

\bibitem{guo2021crossroi}
H.~Guo, S.~Yao, Z.~Yang, Q.~Zhou, and K.~Nahrstedt, ``{CrossRoI: Cross-Camera Region of Interest Optimization for Efficient Real Time Video Analytics at Scale},'' in \emph{Proceedings of the 12th ACM Multimedia Systems Conference}, 2021, pp. 186--199.

\bibitem{yang2023novel}
K.~Yang, J.~Liu, D.~Yang, H.~Wang, P.~Sun, Y.~Zhang, Y.~Liu, and L.~Song, ``A novel efficient multi-view traffic-related object detection framework,'' in \emph{ICASSP 2023-2023 IEEE International Conference on Acoustics, Speech and Signal Processing (ICASSP)}.\hskip 1em plus 0.5em minus 0.4em\relax IEEE, 2023, pp. 1--5.

\bibitem{kohl2020mta}
P.~Kohl, A.~Specker, A.~Schumann, and J.~Beyerer, ``{The mta Dataset for Multi-target Multi-camera Pedestrian Tracking by Weighted Distance Aggregation},'' in \emph{Proceedings of the IEEE/CVF Conference on Computer Vision and Pattern Recognition Workshops}, 2020, pp. 1042--1043.

\bibitem{xu2018deepcache}
M.~Xu, M.~Zhu, Y.~Liu, F.~X. Lin, and X.~Liu, ``{DeepCache: Principled Cache for Mobile Deep Vision},'' in \emph{Proceedings of the 24th Annual International Conference on Mobile Computing and Networking}, 2018, pp. 129--144.

\bibitem{ren2018sbnet}
M.~Ren, A.~Pokrovsky, B.~Yang, and R.~Urtasun, ``{SBNet: Sparse Blocks Betwork for Fast Inference},'' in \emph{Proceedings of the IEEE Conference on Computer Vision and Pattern Recognition}, 2018, pp. 8711--8720.

\bibitem{li2023cross}
J.~Li, L.~Liu, H.~Xu, S.~Wu, and C.~J. Xue, ``{Cross-Camera Inference on the Constrained Edge},'' in \emph{Proc. IEEE INFOCOM}, 2023.

\bibitem{dai2022respire}
X.~Dai, P.~Yang, X.~Zhang, Z.~Dai, and L.~Yu, ``{RESPIRE: Reducing Spatial--Temporal Redundancy for Efficient Edge-Based Industrial Video Analytics},'' \emph{IEEE Transactions on Industrial Informatics}, vol.~18, no.~12, pp. 9324--9334, 2022.

\bibitem{veit2018convolutional}
A.~Veit and S.~Belongie, ``{Convolutional Networks with Adaptive Inference Graphs},'' in \emph{Proceedings of the European Conference on Computer Vision (ECCV)}, 2018, pp. 3--18.

\bibitem{wu2018blockdrop}
Z.~Wu, T.~Nagarajan, A.~Kumar, S.~Rennie, L.~S. Davis, K.~Grauman, and R.~Feris, ``{Blockdrop: Dynamic Inference Paths in Residual Networks},'' in \emph{Proceedings of the IEEE conference on computer vision and pattern recognition}, 2018, pp. 8817--8826.

\bibitem{wang2018skipnet}
X.~Wang, F.~Yu, Z.-Y. Dou, T.~Darrell, and J.~E. Gonzalez, ``{Skipnet: Learning Dynamic Routing in Convolutional Networks},'' in \emph{Proceedings of the European Conference on Computer Vision (ECCV)}, 2018, pp. 409--424.

\bibitem{Chen_2022_CVPR}
Y.~Chen, Y.~Li, X.~Zhang, J.~Sun, and J.~Jia, ``Focal sparse convolutional networks for 3d object detection,'' in \emph{Proceedings of the IEEE/CVF Conference on Computer Vision and Pattern Recognition (CVPR)}, June 2022, pp. 5428--5437.

\bibitem{gao2022convmae}
P.~Gao, T.~Ma, H.~Li, J.~Dai, and Y.~Qiao, ``Convmae: Masked convolution meets masked autoencoders,'' \emph{arXiv preprint arXiv:2205.03892}, 2022.

\bibitem{verelst2020segblocks}
T.~Verelst and T.~Tuytelaars, ``{Segblocks: Towards Block-based Adaptive Resolution Networks for Fast Segmentation},'' in \emph{Computer Vision--ECCV 2020 Workshops: Glasgow, UK, August 23--28, 2020, Proceedings, Part V 16}.\hskip 1em plus 0.5em minus 0.4em\relax Springer, 2020, pp. 18--22.

\bibitem{nalaie2022deepscale}
K.~Nalaie, R.~Xu, and R.~Zheng, ``Deepscale: Online frame size adaptation for multi-object tracking on smart cameras and edge servers,'' in \emph{2022 IEEE/ACM Seventh International Conference on Internet-of-Things Design and Implementation (IoTDI)}.\hskip 1em plus 0.5em minus 0.4em\relax IEEE, 2022, pp. 67--79.

\bibitem{chavdarova2018wildtrack}
T.~Chavdarova, P.~Baqu{\'e}, S.~Bouquet, A.~Maksai, C.~Jose, T.~Bagautdinov, L.~Lettry, P.~Fua, L.~Van~Gool, and F.~Fleuret, ``{Wildtrack: A Multi-camera HD Dataset for Dense Unscripted Pedestrian Detection},'' in \emph{Proceedings of the IEEE conference on computer vision and pattern recognition}, 2018, pp. 5030--5039.

\bibitem{garey1979computers}
M.~R. Garey and D.~S. Johnson, \emph{Computers and intractability}.\hskip 1em plus 0.5em minus 0.4em\relax freeman San Francisco, 1979, vol. 174.

\bibitem{jde}
Z.~Wang, L.~Zheng, Y.~Liu, and S.~Wang, ``{Towards Real-time Multi-object Tracking},'' \emph{arXiv preprint arXiv:1909.12605}, vol.~2, no.~3, p.~4, 2019.

\bibitem{unityEngine}
\BIBentryALTinterwordspacing
``Unity: Unity technologies.'' [Online]. Available: \url{https://unity.com/}
\BIBentrySTDinterwordspacing

\bibitem{reda2022film}
F.~Reda, J.~Kontkanen, E.~Tabellion, D.~Sun, C.~Pantofaru, and B.~Curless, ``Film: Frame interpolation for large motion,'' in \emph{Computer Vision--ECCV 2022: 17th European Conference, Tel Aviv, Israel, October 23--27, 2022, Proceedings, Part VII}.\hskip 1em plus 0.5em minus 0.4em\relax Springer, 2022, pp. 250--266.

\bibitem{CrossRoIimplementation}
``\href{https://github.com/hongpeng-guo/CrossRoI}{{CrossRoI's Implementation}},'' https://github.com/hongpeng-guo/CrossRoI.

\bibitem{2020SciPy-NMeth}
P.~Virtanen, R.~Gommers, T.~E. Oliphant, M.~Haberland, T.~Reddy, D.~Cournapeau, E.~Burovski, P.~Peterson, W.~Weckesser, J.~Bright, S.~J. {van der Walt}, M.~Brett, J.~Wilson, K.~J. Millman, N.~Mayorov, A.~R.~J. Nelson, E.~Jones, R.~Kern, E.~Larson, C.~J. Carey, {\.I}.~Polat, Y.~Feng, E.~W. Moore, J.~{VanderPlas}, D.~Laxalde, J.~Perktold, R.~Cimrman, I.~Henriksen, E.~A. Quintero, C.~R. Harris, A.~M. Archibald, A.~H. Ribeiro, F.~Pedregosa, P.~{van Mulbregt}, and {SciPy 1.0 Contributors}, ``{{SciPy} 1.0: Fundamental Algorithms for Scientific Computing in Python},'' \emph{Nature Methods}, vol.~17, pp. 261--272, 2020.

\end{thebibliography}

\end{document}